\documentclass[runningheads]{llncs}
\usepackage{graphicx}
\graphicspath{ {./figures/} }   

\usepackage{tikz}
\usepackage{comment}
\usepackage{amsmath,amssymb} % define this before the line numbering.
\usepackage{color}
\usepackage{tablefootnote}
\usepackage{hyperref}

\usepackage[accsupp]{axessibility}  % Improves PDF readability for those with disabilities.

\UseRawInputEncoding

\begin{document}
% \linenumbers
\pagestyle{headings}
\mainmatter
\def\ECCVSubNumber{6}  % Insert your submission number here

\title{CrackSeg9k: A Collection and Benchmark for Crack Segmentation Datasets and Frameworks} % Replace with your title

\begin{comment}
\titlerunning{ECCV-22 submission ID \ECCVSubNumber} 
\authorrunning{ECCV-22 submission ID \ECCVSubNumber} 
\author{Anonymous ECCV submission}
\institute{Paper ID \ECCVSubNumber}
\end{comment}
%******************

% \begin{comment}
\titlerunning{CrackSeg9k}
\author{Shreyas Kulkarni\inst{1} \and
Shreyas Singh\inst{1} \and
Dhananjay Balakrishnan\inst{1} \and
Siddharth Sharma\inst{1} \and
Saipraneeth Devunuri\inst{2} \and
Sai Chowdeswara Rao Korlapati\inst{3}}
\authorrunning{S. Kulkarni et al.}
\institute{Indian Institute of Technology, Madras \and
University of Illinois Urbana-Champaign
 \and
Rizzo International, Inc., USA}
% \end{comment}
%******************
\maketitle

\begin{abstract}
The detection of cracks is a crucial task in monitoring structural health and ensuring structural safety. The manual process of crack detection is time-consuming and subjective to the inspectors. Several researchers have tried tackling this problem using traditional Image Processing or learning-based techniques. However, their scope of work is limited to detecting cracks on a single type of surface (walls, pavements, glass, etc.). The metrics used to evaluate these methods are also varied across the literature, making it challenging to compare techniques. This paper addresses these problems by combining previously available datasets and unifying the annotations by tackling the inherent problems within each dataset, such as noise and distortions. We also present a pipeline that combines Image Processing and Deep Learning models. Finally, we benchmark the results of proposed models on these metrics on our new dataset and compare them with state-of-the-art models in the literature.
\end{abstract}

\section{Introduction}

Cracks are common building distress, which is a potential threat to the safety and integrity of the buildings. Localizing and fixing the cracks is a major responsibility in maintaining the building. However, the task of detecting cracks is both tedious and repetitive. To expedite this process and alleviate the workload on experts, it is necessary to achieve automation in crack detection and segmentation.

For a long time, such crack detection has been done manually. Of late, much research has gone into developing automated techniques. Initial work focused on using image thresholding and edge detection techniques ~\cite{fan2019road}~\cite{otsu}~\cite{talab2016detection}~\cite{akagic2018pavement}. Recently, focus has shifted to using Deep Learning for classification~\cite{silva2018concrete}~\cite{ICEM18-05387}~\cite{FLAH2020103781}~\cite{kim2021surface}, object detection~\cite{fan2019road}~\cite{mandal2018automated}~\cite{park2020concrete}~\cite{li2021detection} and semantic segmentation~\cite{yamane2020crack}~\cite{lee2019robust}~\cite{zhang2019concrete}~\cite{shim2020multiscale} of cracks. Attempts have been made to use various techniques, including feature pyramid networks~\cite{yang2019feature} and segmentation models like U-Net~\cite{2021}. Some experiments have been conducted using pre-trained Conv-Nets~\cite{performance}, with limited crack-detection abilities. Better results were obtained when crack segmentation was attempted using models from the R-CNN family~\cite{Kalfarisi2020CrackDA}.

The Deep Learning approaches are highly dependent on the data that has been used for training. The study by \cite{gao2019generative} makes use of 3 datasets: AigleRN~\cite{aiglern} (38 images), CFD dataset~\cite{cfd} (118 images), and the HTR dataset (134 images, not publicly available). Another work by ~\cite{lau2020automated} performs analysis on the CFD dataset~\cite{cfd} and Crack500 dataset~\cite{crack500}.~\cite{oliveira2017road} have evaluated their pipeline on two datasets containing 56 grayscale and 166 RGB images respectively. SDDNet ~\cite{choi2019sddnet} (200 images) manually created a dataset to check their model's performance. DeepCrack~\cite{liu2019deepcrack} too highlights the lack of a benchmark dataset and makes an effort to create a dataset of 537 RGB images, which has been included in our dataset.  

\begin{figure}
    \centering
    \includegraphics[width=\textwidth]{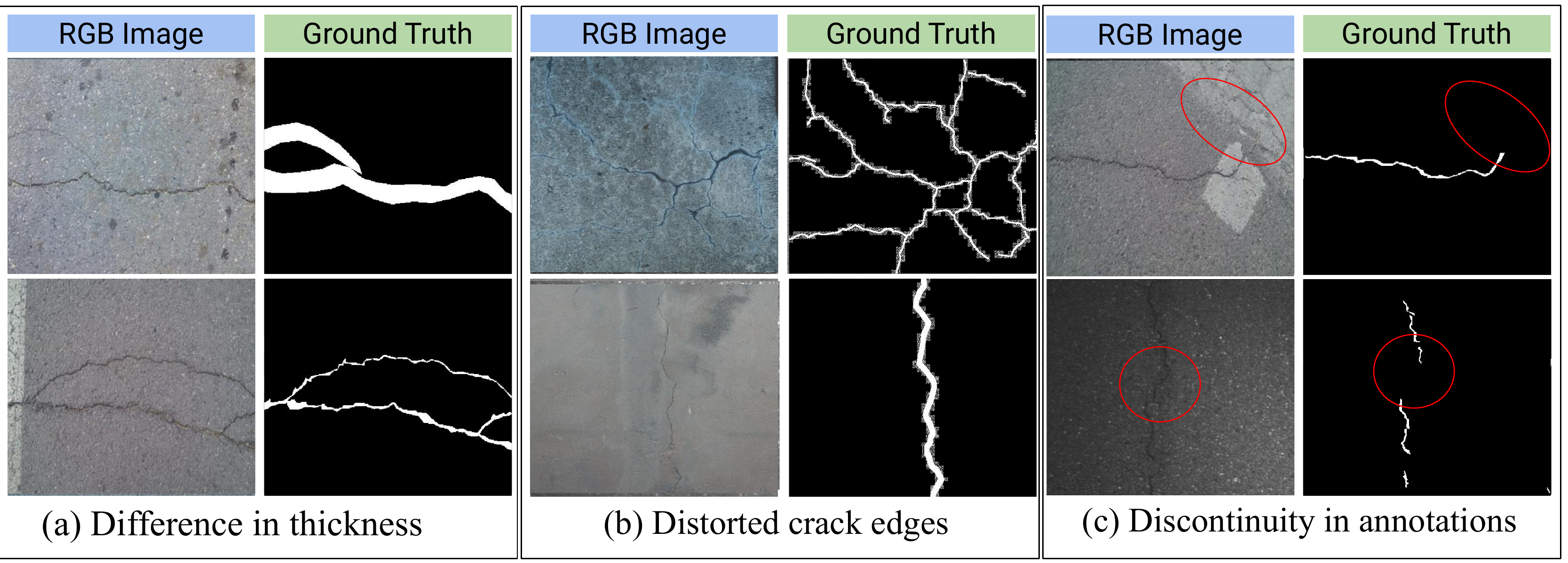}
    \caption{Inconsistent Annotations among sub-datasets}
    \label{fig:inconsistent_crack}
\end{figure}

\vspace{-2mm}
All the attempts mentioned above have fallen short of requirements primarily due to the quality and quantity of publicly available data sets. Obtaining an annotated data set with segmented masks for cracks is a challenge. Most crack data sets have less than 500 images, with several having less than 100 images. Further, the data set being used is not generalizable, and this, combined with the quality of the data sets, leads to non-reproducible results. The datasets used in all cases have different resolutions of the images and different types of annotations leading to inconsistencies as shown in Fig. \ref{fig:inconsistent_crack}. This makes it more difficult to compare with other works. We work towards unifying all the sub-datasets involved to deal with such issues.

We aim to create an efficient Crack segmentation model using the latest Computer Vision and Deep Learning advances. While we did perform bounding box prediction to localize crack regions in images of arbitrary resolutions, our primary focus has remained on binary segmentation of the corresponding cracks. To overcome the limiting factor of the quality of data sets, we combined data sets from various sources while also attempting to unify their ground truths using Image Processing techniques( detailed in Section \ref{subsec:dataset-improv}). This paper makes the following major contributions:

\begin{itemize}
    \item Compilation of a dataset consisting of 9000+ images of cracks on diverse surfaces with annotations for semantic segmentation. The compiled data is also categorized into linear, branched, and webbed.
    \item Further refinement of the ground truth masks by denoising and unifying the annotations using Image Processing techniques per the dataset's requirements. 
    \item An end-to-end pipeline has been devised for efficient localization and segmentation of cracks for images of arbitrary lighting, orientation, resolution and background. 
    \item We explore various rule-based and data-driven techniques for crack segmentation to establish a benchmark on our dataset.
    \item Finally, we present an approach to fuse unsupervised self-attention mechanisms into CNNs for improvising over the SOTA models.

\end{itemize}

\section{Related Works}

\subsection{Rule based Image Processing Techniques}
\label{subsec:rule-based}

Rule-based Image Processing methods have been traditionally used for image-based crack segmentation. The techniques can broadly be classified into two categories, i.e., Edge detection and Thresholding \cite{KHERADMANDI2022126162}.

Thresholding complies with the rule of classifying the pixels of a given image into groups of objects and background; the former includes those pixels with gray values more than a specific threshold, and the latter comprises those with gray values equal to and below the threshold. Choosing an appropriate threshold value is the main challenge in this method. Using a histogram to choose a threshold value gives satisfactory results for segmenting images of road surfaces~\cite{mahler1991pavement}. 

 The purpose of the edge-based algorithm is to segment an image based on the discontinuity detection technique~\cite{sonka1999segmentation}. The focus of edge-based algorithms is on the linear features that mainly correspond to crack boundaries and interesting object surface details. Based on the comparative study, the algorithms calculate the magnitude for each pixel and use double Thresholding to discover potential edges. Even though useful for edge detection, these algorithms can only detect edges and cannot extrapolate over the entire crack surface; this makes edge detection ineffective for binary segmentation.

% For our research, rule-based algorithms have been used as stand-alone techniques for segmenting cracks and extracting features from images that can be integrated into the pipeline.

\subsection{Data Driven Methods}
\label{subsec:data-driven}

The lack of generalizability of rule-based methods to diverse environments with different backgrounds and lighting has promoted research into the data-driven detection of cracks. These techniques comprise of data-hungry Deep Learning models that are either fully convolutional, attention based or a mix of both, requiring the training of millions of parameters.

While traditional object detection algorithms re-purposed classifiers to perform detection, \textbf{YOLO}\cite{7780460} framed object detection as a regression problem to spatially separated bounding boxes and associated class probabilities. A single neural network directly predicts bounding boxes and class probabilities from full images in one evaluation. We have trained a YOLO-v5 object detector to localize cracks in images of arbitrary size to crop them before pixel-wise segmentation.

The image-to-Image translation is the spatial transformation of images from one domain to another. This field has grown immensely since \textbf{Pix2Pix GAN}~\cite{isola2018imagetoimage} was introduced. GANs~\cite{gans} consist of two parts: a generative model $G$ that captures the data distribution and a discriminative model $D$ that estimates the probability that a sample came from the training data rather than $G$. The training procedure for $G$ is to maximize the probability of $D$ making a mistake. We pose semantic segmentation of cracks using Pix2Pix as an image-to-image translation problem, where the generator translates an input RGB image into a binary segmentation mask for cracks, and the discriminator tries to distinguish b/w real and fake samples conditioned on the input. The \textbf{U-Net}\cite{u-net} architecture is used for semantic segmentation, which consists of a contracting path to capture context and a symmetric expanding path that enables precise localization. This has been utilized as a backbone for the Generator of GANs.

\textbf{DeepLab}\cite{deeplab} highlights convolution with upsampled filters, or `atrous convolution' and ASPP `Atrous Spatial Pyramidal Pooling', as a powerful tool in dense segmentation tasks. DMA-Net~\cite{dma-net} proposes a multi-scale attention module in the decoder of DeepLabv3+ to generate an attention mask and dynamically assign weights between high-level and low-level feature maps for crack segmentation. 

\textbf{Swin-Unet}\cite{swin-unet} is an U-Net-like pure Transformer for image segmentation. The tokenized image patches are fed into the Transformer-based U-shaped Encoder-Decoder architecture with skip-connections for local-global semantic feature learning. Specifically, a hierarchical Swin Transformer with shifted windows is used as the encoder to extract context features. Moreover, a symmetric Swin Transformer-based decoder with a patch expanding layer is designed to perform the up-sampling operation to restore the spatial resolution of the feature maps. 

\textbf{DINO}~\cite{dino} highlights that self-supervised Vision Transformer features contain explicit information about the semantic segmentation of an image, which does not emerge as clearly with supervised Vision Transformers, nor with Convolutional Nets. Self-supervised ViT features explicitly contain the scene layout and, in particular, object boundaries. This information is directly accessible in the self-attention modules of the last block.

\section{Dataset}

The proposed dataset is the largest, most diverse and consistent crack segmentation dataset constructed so far. It contains a total of 9255 images, combining different smaller open source datasets. Therefore, the dataset has great diversity in the surface, background, lighting, exposure, crack width and type (linear, branched, webbed, non-crack). Smaller datasets like Eugen Muller~\cite{eugen-muller} and Sylvie Chambon~\cite{aiglern} were discarded due to poor quality and much fewer images. The DIC dataset~\cite{dic-dataset} containing 400+ images was avoided because the images were augmented from a tiny set of 17 images and had very little variety. The Forest dataset \cite{shi2016automatic} was excluded as it was found to be very similar to the CFD dataset.

This dataset is divided into training, validation and testing sets with 90\%, 5\% and 5\% split. The final dataset is made available to public on Harvard Dataverse\cite{crackseg9k}. 

\subsection{Dataset Details}

Table \ref{table:subdataset} summarises the sub-datasets involved. This subsection will discuss the sub-datasets involved by providing an overview of data collection techniques, characteristics of image and dataset size.

\setlength{\tabcolsep}{4pt}
\begin{table}
\begin{center}
\caption{Sub dataset quantitative details and performance on benchmark models}
\label{table:subdataset}
\resizebox{\columnwidth}{!}{%
\begin{tabular}{ |l|c|c|c|c|c|c|c| } 
\hline\noalign{\smallskip}
\textbf{ Dataset} & \textbf{Size} & \textbf{Resolution} & \textbf{\% of} & \multicolumn{2}{|c|}{\textbf{Pix2Pix}} & \multicolumn{2}{|c|}{\textbf{Deeplab}} \\ 
 & & & \textbf{cracks} &  \multicolumn{2}{|c|}{\textbf{U-Net}} & \multicolumn{2}{|c|}{\textbf{ResNet *}} \\
 & & & &  \textbf{mIoU}  & \textbf{F1 } & \textbf{mIoU}  & \textbf{F1}\\
\noalign{\smallskip}
\hline
\noalign{\smallskip}
 Masonry & 240 & 224 $\times$ 224& 4.21 & 0.4685 & 0.0392 & 0.4986 & 0.0420\\
 \hline
CFD & 118 & 480 $\times$ 320 & 1.61 & 0.6100 & 0.3942 & 0.6232 & 0.4203\\  
 \hline
  CrackTree200 & 175 & 800 $\times$ 600 & 0.31 & 0.5030 & 0.0799 & 0.5478 & 0.1132\\
 \hline
 Volker & 427 & 512 $\times$ 512 & 4.05 & 0.6743 & 0.5423 & 0.8209 & 0.7955 \\ 
 \hline
 DeepCrack & 443 & 544 $\times$384 & 3.58 & 0.7207 & 0.6193 & 0.8311 & 0.8068 \\
 \hline
 Ceramic & 100 & 256 $\times$ 256& 2.05 & 0.4783 & 0.0330 & 0.5095 & 0.0480\\
 \hline
SDNET2018 & 1411 & 4032 $\times$ 3024 & 0 & 0.4865 & N/A & 0.5000 & N/A\\
\hline
 Rissbilder & 2736 & 512 $\times$ 512 & 2.70 & 0.6050 & 0.3854 & 0.7512 & 0.6856  \\
 \hline
Crack500 & 3126 & 2000 $\times$ 1500 & 6.03 & 0.6495 & 0.4974 & 0.8230 & 0.8032 \\ 
  \hline
GAPS384 & 383 & 640 $\times$ 540 & 1.21 & 0.5716 & 0.2825 & 0.5965 & 0.3358 \\ 
 \hline

\end{tabular}%
}
\end{center}
\end{table}
\setlength{\tabcolsep}{1.4pt}

The \textbf{Masonry}\cite{masonry} dataset contains images of masonry walls. They consist of images from the Internet and masonry buildings in the Groningen, Netherlands. The crack patches comprise of small (a couple of bricks) to larger (whole masonry walls) fields of view.  

\textbf{CFD}\cite{cfd} dataset is composed of images reflecting urban road surface conditions in Beijing, China. Each image has hand-labelled ground truth contours. \textbf{CrackTree200}\cite{cracktree200}, \textbf{Volker}\cite{wall-climbing}   and \textbf{DeepCrack}\cite{liu2019deepcrack} contain images with manually annotated cracks collected from pavements and buildings. They have challenges like shadows, occlusions, low contrast and noise.  

\textbf{Ceramic Cracks Database}\cite{ceramic} was collected by students of the University of Pernambuco. The database has images of building facades with cracks on different colours and textures of ceramics.

\begin{figure}[h]
    \centering
    \includegraphics[width=\textwidth]{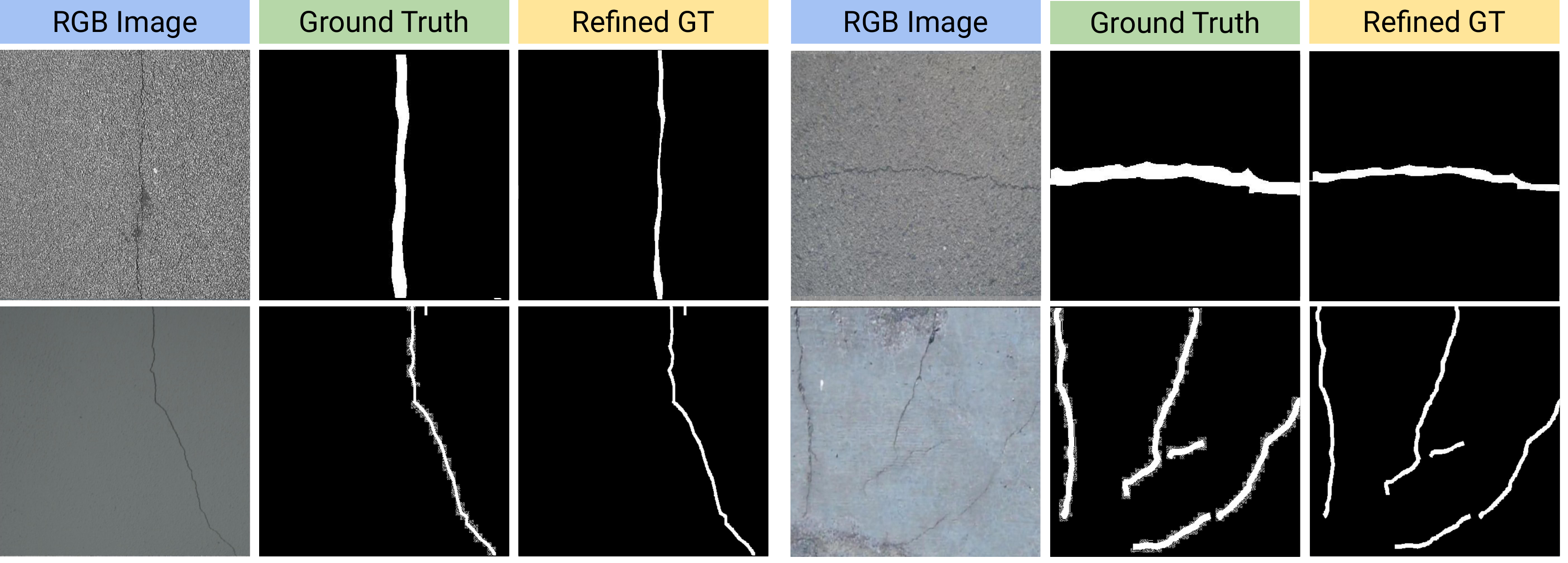}
    \caption{Refined Dataset. Hairline distortions have been tackled in the lower half of the images while the thickness has been reduced in the upper half of the images}
    \label{fig:dataset-refined}
\end{figure}

\textbf{SDNET2018}\cite{sdnet2018} contains various non-crack images from concrete bridge decks, walls, and pavements. The dataset includes cracks as narrow as 0.06 mm and as wide as 25 mm. The dataset also includes images with various obstructions, including shadows, surface roughness, scaling, edges, holes, and background debris.

\textbf{Rissbilder}\cite{wall-climbing} consists of varied types of architectural cracks, and \textbf{Crack500}\cite{crack500} was collected at the Temple University campus using a smartphone as the data sensor. The \textbf{GAPS}\cite{gaps384} data acquisition occurred in the summer of 2015, so the measuring condition was dry and warm.

\subsection{Dataset Refinement}
\label{subsec:dataset-improv}
% write about resolution

\setlength{\tabcolsep}{4pt}
\begin{table}[h]
\begin{center}
\caption{Sub dataset Refinement Summary}
\label{table:refinement-summary}
\begin{tabular}{ |l|l|l| } 
\hline\noalign{\smallskip}
 \textbf{Dataset} & \textbf{Distortion} & \textbf{Refinement} \\ 
\noalign{\smallskip}
\hline
\noalign{\smallskip}
Masonry   & None & None \\ 
 \hline
CFD   & None & None \\ 
 \hline
  CrackTree & Continuity errors; & Dilation followed by  \\ 200 & Hairline distortion & Erosion with special kernels \\
 \hline
Volker & Randomly thick cracks; & Contour area based erosion; \\ 
 & Hairline distortion & Morph opening \\ 
 \hline
 DeepCrack & Crack boundary irregular  & None \\
 \hline

Ceramic   & None & None \\ 
 \hline
 SDNET2018 & None & None \\
 \hline
 Rissbilder & Hairline distortion & Morph opening with  \\ for Florian &  & increased kernel size \\
 \hline 
Crack500 &  Extra thick and & Erosion and \\ 
& hairline distortion & morph opening \\ 
 \hline
 GAPS384 & Random thickening & Erosion or Dilation on \\ 
 & of cracks & the basis of Contour size \\ 
 \hline

\end{tabular}
\end{center}
\end{table}
\setlength{\tabcolsep}{1.4pt}

The open source datasets had different resolutions and the ground truths had many distortions and discontinuities. They were also vastly inconsistent in their thickness(see Fig.\ref{fig:inconsistent_crack}). Many cracks had a hairlike boundary noise surrounding them. This led to scenarios where the model's metric score was less even for visually accurate predictions due to the distortions in the Ground Truths.

Firstly, we resize all the images to a standard size of 400$\times$400. For images greater than this size, random crops of this size were extracted from images, and images with resolution less than it were linearly interpolated to the standard size. Once the uniformity in size was attained, we performed Gaussian blurring followed by morphological operations to remove noise from the image. A 3x3 kernel made up of ones was used to erode the mask, followed by dilation with the hairline distortions to remove it. Depending on the distortion (thickened or thinned), the images required either erosion or dilation. We used the contour area as the metric for finding the kind of distortion and operating on it, respectively. We have summarised all the transformations performed on the sub-datasets in the Table \ref{table:refinement-summary} and sample refinements can be visualised in Fig.\ref{fig:dataset-refined}.

\subsection{Categorizing Dataset based on Crack Types}
\label{subsec:crack-type}
The dataset has been divided into three subtypes, namely Linear, Branched and Webbed, based on the spatial patterns of cracks observed in the images. To achieve the classification for the entire dataset, we trained a small ResNet-18 pre-trained classifier on 150 images, with 50 belonging to each class. We manually labelled these 150 images to generate ground truths for classification. This methodology generalized well to our entire dataset. 
\begin{figure}[!htp]
    \centering
    \includegraphics[width=\textwidth]{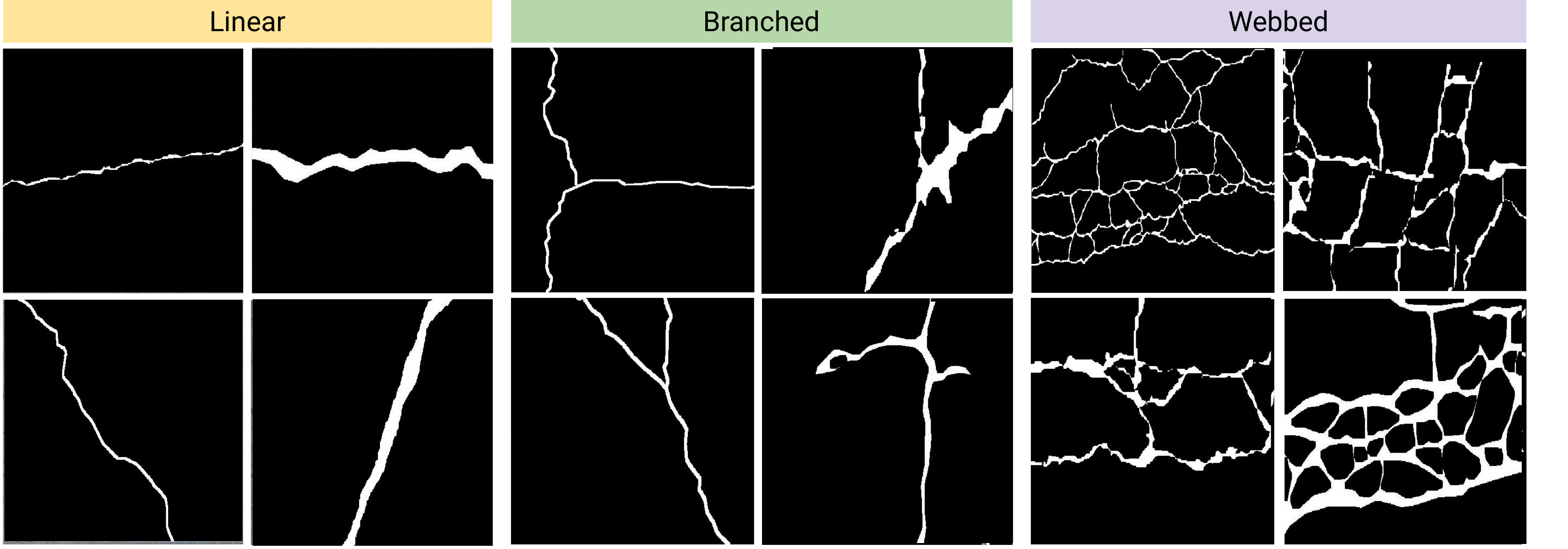}
    \caption{Sample ground truths belonging to each category of crack type}
    \label{fig:types}
\end{figure}
The cracks with no branches are classified as linear, Cracks with 2-5 branches are classified as branched, and cracks with more than five cracks are classified as webbed. This division of the dataset aims to benchmark and contrast the generalizing ability of benchmarked models across various crack modalities and complexity for detection. Fig. \ref{fig:types} shows the cracks belonging to different classes, and Table \ref{table:types} contains details of the crack types and their distribution in the dataset.
\setlength{\tabcolsep}{4pt}
\begin{table}
\begin{center}
\caption{performance based on the type of crack}
\label{table:types}
\begin{tabular}{ |c|c|c|c|c|c|c|c| } 
\hline\noalign{\smallskip}
 \textbf{Dataset} & \textbf{Size} & \textbf{Percentage} & \multicolumn{2}{|c|}{\textbf{Pix2Pix}} & \multicolumn{2}{|c|}{\textbf{Deeplab}} \\ 
 & & \textbf{of cracks} &  \multicolumn{2}{|c|}{\textbf{U-Net}} & \multicolumn{2}{|c|}{\textbf{ResNet *}} \\
 & & &  \textbf{mIoU}  & \textbf{F1 } & \textbf{mIoU}  & \textbf{F1}\\
\noalign{\smallskip}
\hline
\noalign{\smallskip}
Linear & 2369 & 4.94 & 0.6687 & 0.5455 & 0.8219 & 0.7991\\  
 \hline
Branched & 4192 & 5.45 & 0.6184 & 0.4374 & 0.7940 & 0.7643\\ 
 \hline
Webbed & 1283 & 9.81 & 0.5963 & 0.4259 & 0.6488 & 0.5835\\
 \hline
Non-crack & 1411 & 0 & 0.4865 & N/A & 0.5000 & N/A\\
 \hline

\end{tabular}
\end{center}
\end{table}

\section{Methodology \& Experiments}
\label{sec:print}

This section will discuss the experiments conducted to benchmark and evaluate different models for crack segmentation. These experiments included an amalgamation of both supervised and unsupervised techniques. The experiments were performed on the refined dataset with a standard resolution of 400 $\times$ 400 for all images. 

\subsection{Pipeline}
\begin{figure}[!h]
    \centering
    \includegraphics[scale = 0.3]{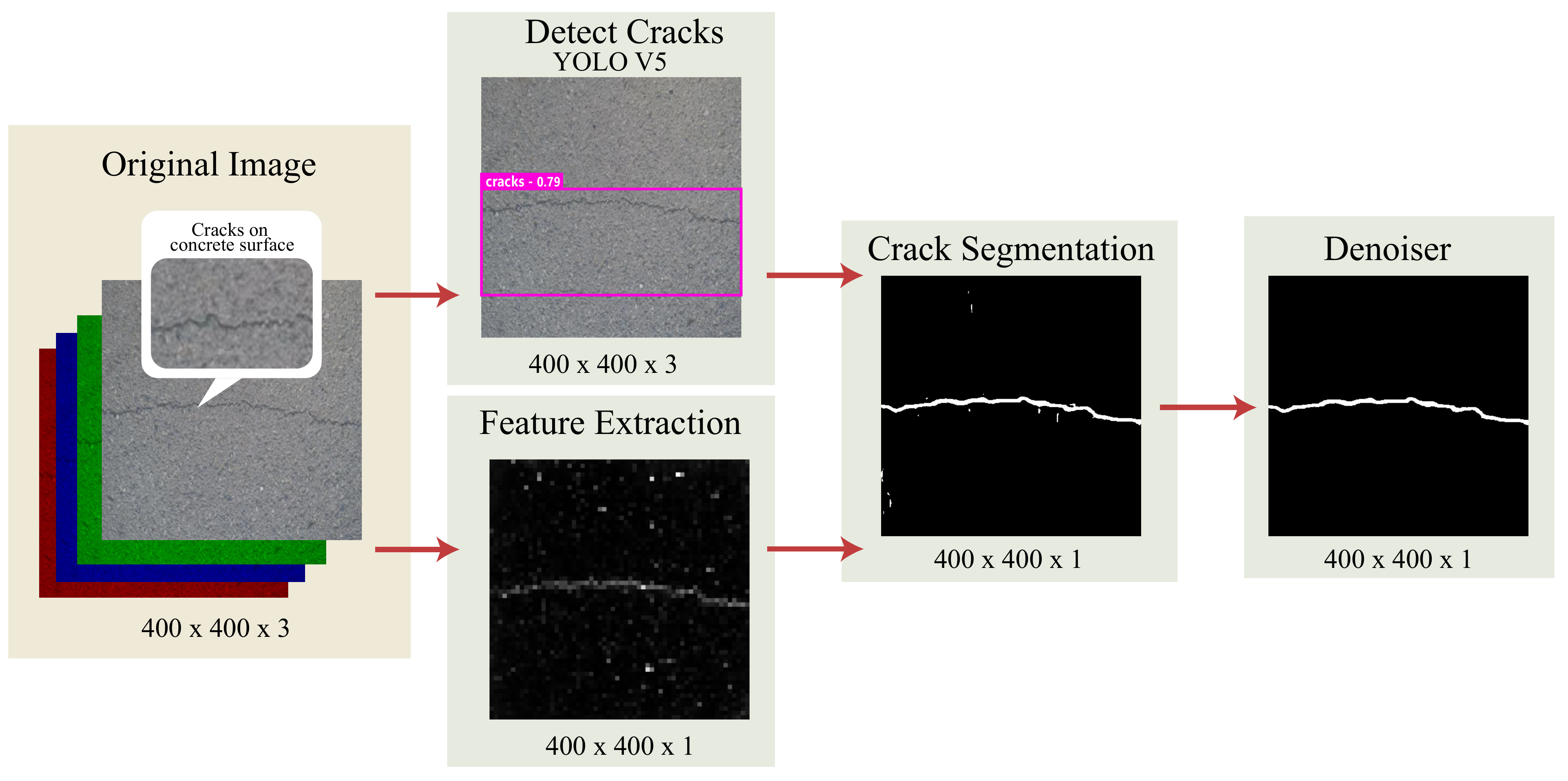}
    \caption{General Pipeline}
    \label{fig:pipeline}
\end{figure}

This subsection explains the details of our pipeline for end-to-end crack segmentation. The pipeline processes the input image through four steps (see Figure \ref{fig:pipeline}). The first step consists of a YOLO-v5-based object detector that localizes the cracks in an input image of arbitrary resolution and draws bounding boxes for the cracks in the image. We create bounding box labels for object detection enclosing the crack using the extreme pixels of the segmented masks (ground truths). In the second step, cracks are cropped using the bounding box coordinates obtained in the previous step and resized to 400$\times$400. The resized crack images are passed through the DINO model to extract unsupervised features that will be used as the prior for the segmentation model. 

\begin{figure}[!h]
    \centering
    \includegraphics[width=\textwidth]{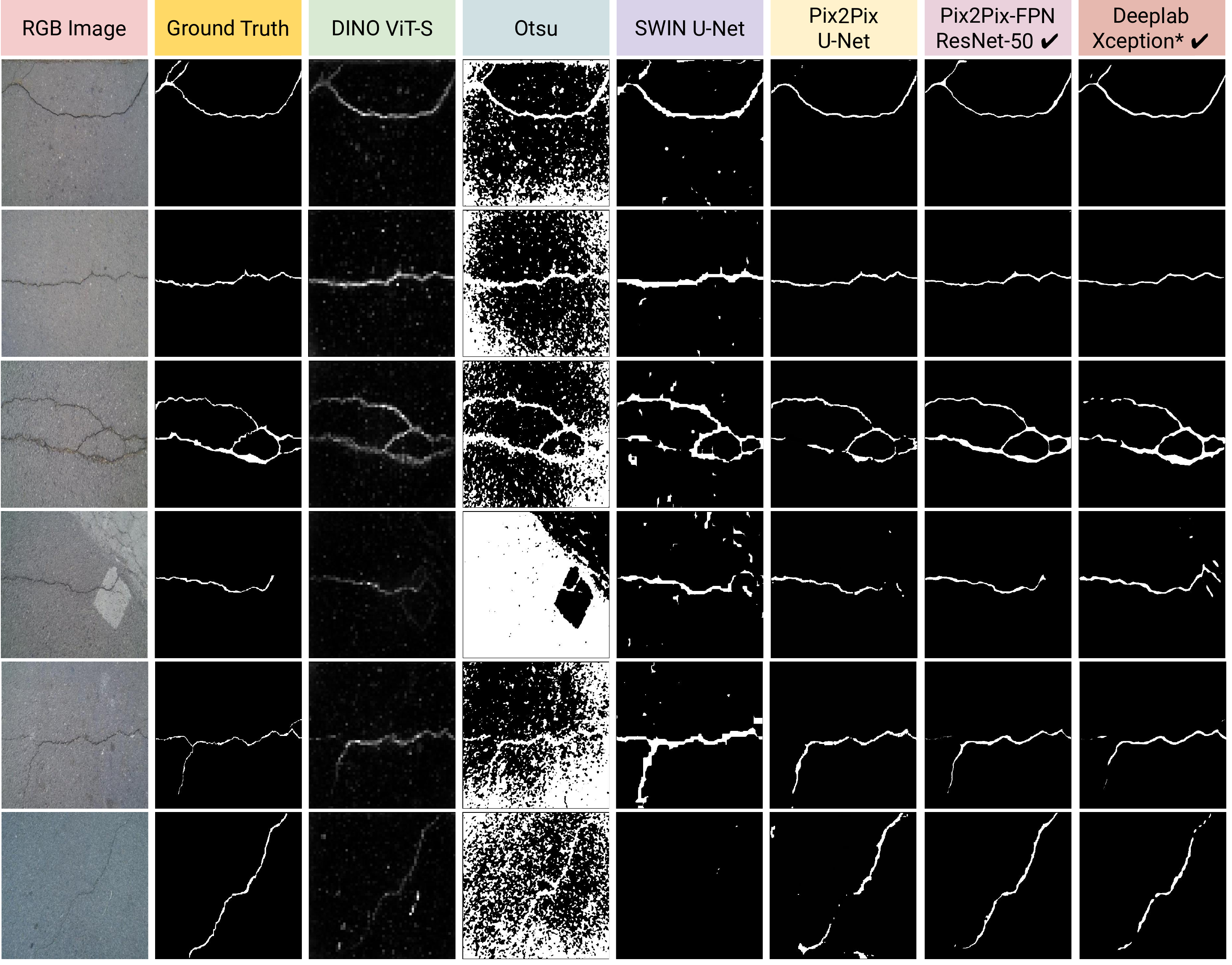}
    \caption{Benchmarked Model Performance Comparisons on Refined Dataset}
    \label{fig:benchmark}
\end{figure}

In the final step, the unsupervised features are concatenated with the resized crack image and fed into an FPN-based binary segmentation model to get pixel-wise classification for the cracks in each image. The final post-processing step passes the segmentation maps through a denoiser; the denoised segmentation mask is reshaped to the size of its original bounding box. Finally, all the predictions for a given region are stitched together to produce the final output. The results on various techniques have been benchmarked in Table \ref{table:performance-summary} and images with their predictions on the respective models are showcased in Fig.\ref{fig:benchmark}. Note: the purpose of the YOLO object detector is to concentrate the segmentation model's focus on a patch of cracks in case of high-resolution images. 

% use Adobe illustrator and save as svg or pdf and make it better!

\setlength{\tabcolsep}{4pt}
\begin{table}
\begin{center}
\caption{
Model Performance Summary \tablefootnote{* - pre-training on ImageNet. 
$\checkmark$ - DINO features used as prior for training}}
\label{table:performance-summary}
\begin{tabular}{ |lc|c|c|c| } 
\hline\noalign{\smallskip}
 \textbf{Technique } & \textbf{Backbone} & \textbf{Trainable} &\textbf{F1 Score} & \textbf{MIoU} \\ 
 & & \textbf{Parameters} & &  \\ 
 & & \textbf{in Millions} & &  \\ 

 \hline
 Otsu & None & 0 & 0.0609 & 0.3187 \\ 
 \hline
 DINO & ViT-S & 0 & 0.1568 & 0.5149 \\
 
 \hline
 MaskRCNN & ResNet-50 & 44.178 & 0.4761 & 0.5213 \\ 
 \hline
 SWIN Transformer & U-Net & 41.380 & 0.5009 & 0.6426 \\
 \hline
 Pix2Pix & U-Net & 54.421 & 0.5652 & 0.6666 \\
 \hline
 Pix2Pix $\checkmark$ & U-Net & 54.422 & 0.5824 & 0.6953 \\
 \hline

 Pix2Pix & FPN-ResNet-50 & 26.911 & 0.3950 & 0.6286 \\
 \hline
 Pix2Pix $\checkmark$ & FPN-ResNet-50 & 26.914 & 0.5765 & 0.7142 \\
 \hline
 DeepLab V3+ & Xception * & 54.700 & 0.6344 & 0.7208 \\ 
 \hline
 DeepLab V3+$\checkmark$ & Xception *  & 54.700 & 0.6608 & 0.7353 \\ 
 \hline
 DeepLab V3+& ResNet-101 *  & 59.339 & 0.7060 & 0.7599 \\ 
 \hline
 DeepLab V3+$\checkmark$ & ResNet-101  *  & 59.342 & 0.7238 & 0.7712\\ 
 \hline
\end{tabular}
\end{center}
\end{table}
\setlength{\tabcolsep}{1.4pt}

\subsection{Unsupervised Feature Extraction}
In our first set of experiments, we attempt to generate the predicted masks without utilizing ground truth annotations.

\textbf{Otsu Thresholding} method failed to find the optimum threshold for images with noise. When the dimensions of the cracks were small relative to the background, the technique failed to identify the cracks. Otsu's thresholding method was also unsuccessful in effectively segmenting cracks when the images were taken under non-uniform lighting conditions.

\textbf{DINO for ViT} We used the ViT-Small model trained in a fully unsupervised fashion using fully self-supervised DINO~\cite{dino} which generates self-attention maps for the input images. The generated maps were found to be very noisy, but the self-attention maps captured the branched cracks very successfully due to the global self-attention mechanism of Vision Transformers. Unlike Vision Transformers, CNN-based algorithms lack the global field of view and are dataset antagonistic due to weight sharing, due to which CNN-based algorithms underperform on branched and webbed cracks. The global attention maps generated by transformers via self-supervision will be further utilized in Section \ref{subsec:dino-prior} to generate priors for CNN-based segmentation algorithms to mitigate the problems mentioned above. 

\subsection{Supervised Techniques} 

The unsupervised techniques gave noisy results and failed to capture the domain essence. Therefore, these techniques could not be used directly to infer results and merely act as priors for supervised techniques. In an attempt to improve the results, we conducted experiments using supervised techniques. Supervised techniques involve the utilization of segmented ground truths and they are more robust than unsupervised methods for this reason. We used four supervised Deep Learning algorithms for Image segmentation: MaskRCNN, DeepLab, Pix2Pix and Swin-Unet.

We trained \textbf{MaskRCNN} model using a ResNet-50~\cite{resnet} backbone. The results obtained are significantly better than that of unsupervised techniques. However, we noticed two major issues with Mask-RCNN, the model does not generalize well to the thickness of the cracks despite the diversity in the dataset and is very susceptible to background noise such as debris and shadows, which led to a lot of false positives.

 \textbf{Swin-Unet}~\cite{swin-unet} model was used on our dataset to measure the performance of transformer-based backbones trained end to end in a supervised fashion. The Swin Transformer blocks can model the long-range dependencies of different parts in a crack image from a local and global perspective, which leads to superior performance in the detection of branched and webbed crack.
 
 We conducted two different experiments with \textbf{Pix2Pix} model, using Vanilla U-Net with skip connections and Feature Pyramidal Network with a resnet-50 backbone as the generators. The Patch GAN discriminator used in the Pix2Pix model improves the quality of predictions significantly. The models are able to detect finer cracks and the boundaries of cracks with higher precision. The experiments also highlight that Feature pyramidal networks are superior for segmentation tasks than U-Nets, as they achieve higher accuracy with less than half the number of parameters of U-Nets.
 
 \textbf{DeepLab} V3+ using  ResNet-101 and Xception~\cite{xception} backbones pre-trained on imagenet as encoders. DeepLab is able to detect very fine-grained cracks even in adverse lighting conditions, the predicted masks generalize well to the thickness of the cracks, and in some cases, the model was able to detect cracks that were not annotated in the ground truth. The qualitative and quantitative results indicate DeepLab as the best model among our experiments.

   \begin{figure}[h]
    \centering
    \includegraphics[width=\textwidth]{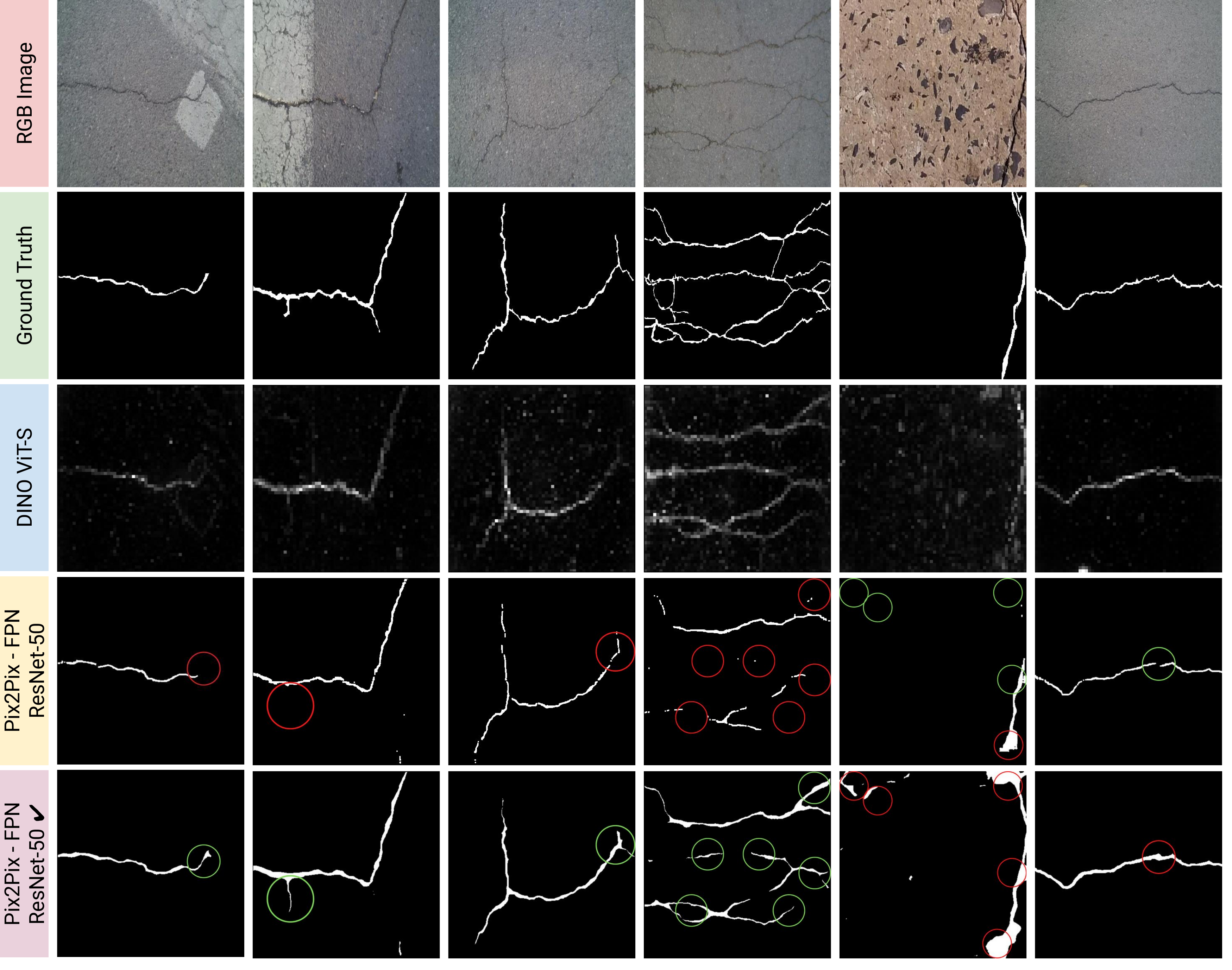}
    \caption{Transformation in predictions when features are integrated}
    \label{fig:features}
\end{figure}

\subsection{Using Semi-Supervised DINO features as prior for segmentation}
 \label{subsec:dino-prior}
 We conducted this experiment to verify our hypothesis that using semi-supervised DINO features prior to CNN-based image segmentation models leads to superior results on branched and webbed cracks. This is because global and local self-attention from transformers helps in establishing long-range dependencies in images and CNNs help in learning dataset-specific localized features.
 For this experiment, we concatenated the semi-supervised self-attention map generated from DINO as the fourth channel to our input image tensor. We conducted this experiment with all our CNN-based models and observed the following: FPN Pix2Pix(14\% mIoU increase), U-Net Pix2Pix(4\% mIoU increase), Resnet-101 DeepLab (2\% mIoU increase) and Xception DeepLab (1\% mIoU increase). To highlight the improvement in performance due to features, please refer Fig. \ref{fig:features}. While integrating features definitely showed significant improvement in the predictions, it  did have a couple of issues. Sometimes, the prediction was affected by the presence of noise in the background, resulting in an erroneous prediction, and at other times, the predicted crack turned out to be thicker than the actual, resulting in the loss of some fine details. These shortcomings have been highlighted in the last 2 columns of \ref{fig:features}.
   
 These experiments demonstrate that adding prior information to the input regarding the localization of cracks and scaling the network compute enough to process the additional information leads to a subsequent increase in model performance.

\subsection{Comparative Analysis of Results}

\begin{figure}[h]
    \centering
    \includegraphics[width=\textwidth]{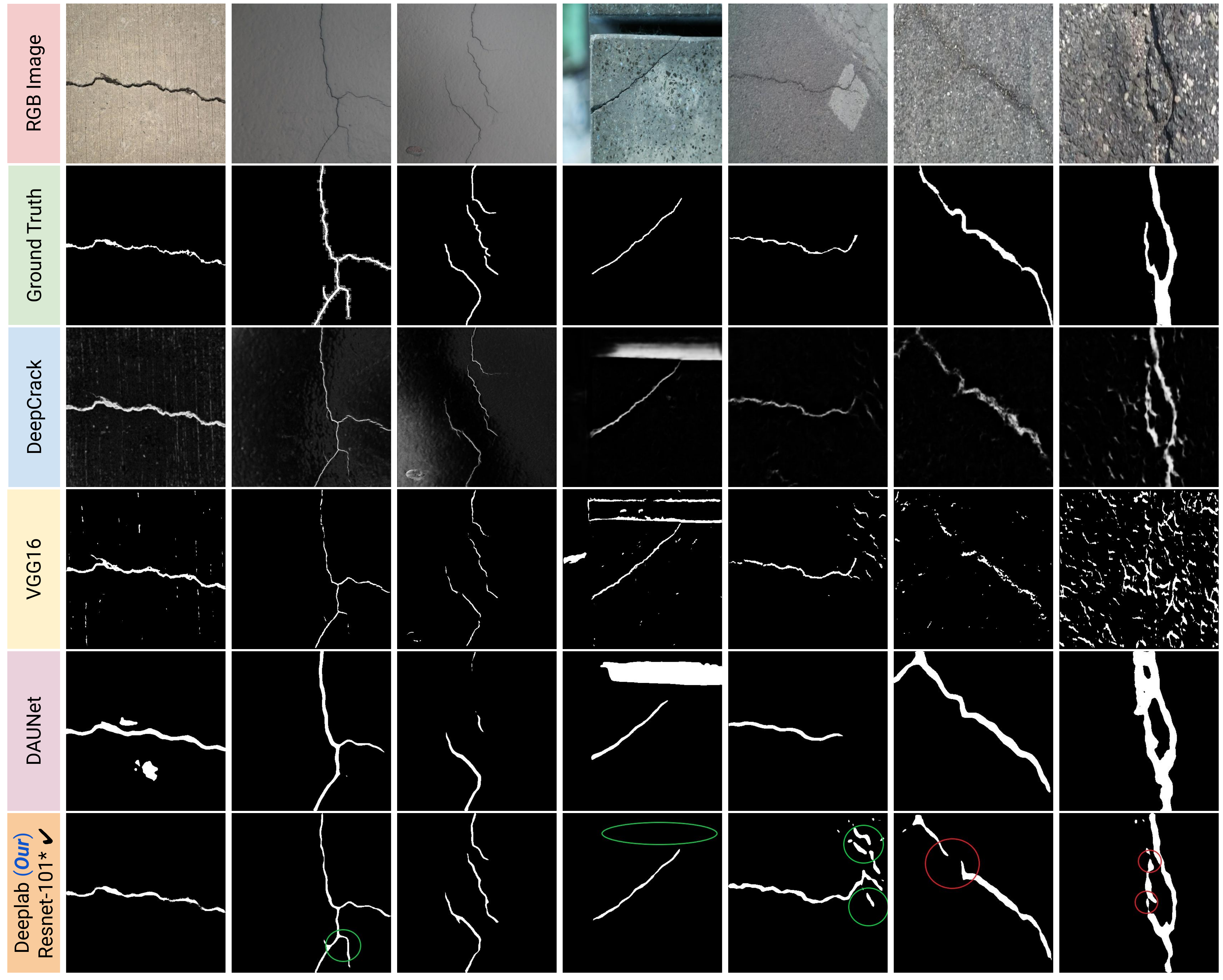}
    \caption{Comparison with state-of-the-art models}
    \label{fig:SOTA}
\end{figure}

\setlength{\tabcolsep}{4pt}
\begin{table}
\begin{center}
\caption{
Comparison with previous works evaluated on our dataset}
\label{table:SOTA}
\begin{tabular}{ |l|c|c| } 
\hline\noalign{\smallskip}
 \textbf{Technique}  &\textbf{F1 Score} & \textbf{MIoU} \\ 
 \hline
 VGG16~\cite{vgg} & 0.3009 & 0.5444\\ 
 \hline
 DeepCrack~\cite{liu2019deepcrack}  & 0.3404 & 0.5813\\ 
 \hline
 DAU-Net~\cite{dau-net} & 0.5495 & 0.6857\\ 
 \hline

DeepLab $\checkmark$ ResNet * (ours)  & 0.7238 & 0.7712 \\
\hline
 
\end{tabular}
\end{center}
\end{table}
\setlength{\tabcolsep}{1.4pt}

We performed a systematic quantitative (Table \ref{table:SOTA}) and qualitative (Fig. \ref{fig:SOTA}) analysis of existing SOTA models for crack segmentation in the literature and compared their performances to the proposed models. All the models were evaluated on the test split of the refined dataset. We observed that DeepCrack scales well to different sizes of cracks due to the aggregation of multi-scale feature maps, but it does not accurately adapt to the varying thickness of cracks and is prone to a lot of noise in the output. DAU-Net and VGG16 do not perform well on untrained datasets; this highlights their inability to generalize across different crack surfaces and illumination conditions, which is largely attributed to the lack of diversity in their training data. However, our model is not perfect. As seen in the diagram, the bad predictions are marked with red circles. We observed that our model sometimes tended to have predictions with small discontinuities and gaps, resulting in the loss of crack information.

Analysis of results on different subsections of our dataset is presented in Table \ref{table:subdataset} and Table \ref{table:types}; we can observe that even though our dataset contains diverse crack images, the benchmarked models perform better for sub-datasets Volker, DeepCrack and Crack500 because they contain a larger number of images. Masonry and Ceramic datasets had very different background textures from the other images, which caused a downgrade in the model's performance. Analysis of the performance of benchmarked models based on the spatial types of cracks highlighted that, on average Linear cracks had the best predictions, followed by branched cracks, and the webbed cracks had the lowest quality of predictions. 

\section{Conclusion}
Within the past decade, Computer Vision techniques for crack detection have gained significant momentum and led to stimulating works and datasets. However, most works focus on a particular dataset and do not benchmark the results on other datasets, all while using varying metrics. This leads to problems in reproducibility and/or limitation in application to different types of cracks such as asphalt, concrete, masonry, ceramic and glass. 

This paper tries to address the aforementioned vital issues that are pertinent. We do so by compiling various datasets and putting together a robust dataset consisting of 9000+ images with varying backgrounds, types of cracks and surfaces, and ground truth annotations. Further, we refined existing datasets and standardized the dataset by maintaining consistency in the image resolution. We further divide the dataset on the basis of the spatial arrangement of cracks into linear, branched and webbed. Finally, we benchmarked the state-of-the-art techniques on our final dataset. 

Our results show that DeepLab with ResNet and XceptionNet as backbone perform the best. We further observe that the models perform best on images with linear cracks, and predictions' quality decreases for webbed and branched cracks. The paper highlights the advantages of using self-supervised attention in modelling long-range dependencies in vision tasks and presents a method to fuse semi-supervised attention feature maps with CNNs for enhanced crack detection. Adding the DINO features as prior results in significant improvements in the predictions with a marginal tradeoff in the number of trainable parameters. % mention 

%   Limitations..
Despite our best efforts in compiling a dataset containing cracks in different domains, the models trained by us have not been able to identify cracks well in the case of transparent backgrounds like glass. We believe in the need for more images belonging to such backgrounds so that models trained on them shall be more domain invariant. Further exploration into fusing semi-supervised and supervised training paradigms and better amalgamation of transformer and CNN architectures can be explored for better performance on images with branched and webbed cracks. 

\bibliographystyle{splncs04}
\bibliography{038}

\appendix

\section{Experimental Settings}
\label{sec:experimental-settings}
For all our experiments, we use a stratified train test split of 90-10\% on the compiled dataset. All the benchmarked models have been trained on Nvidia-GeForce 1080 Ti with 12 GB of Memory.Early Stopping with a patience of 5 epochs was used to determine convergence for all the models.
\setlength{\tabcolsep}{4pt}
\begin{table}[h]
\begin{center}
\caption{Model Hyperparameters for reproducibility}
\label{table:refinement-summary}
\begin{tabular}{ |l|l|l|l|l| } 
\hline\noalign{\smallskip}
 \textbf{Model} & \textbf{Optimizer \&} & \textbf{Epochs}  & \textbf{Pre-Trained} & \textbf{Encoder} \\ 
 & \textbf{Learning Rate} &   &  &  \\ 

\noalign{\smallskip}
\hline
\noalign{\smallskip}
MaskRCNN & Adam , 1e-4 & 20 & ImageNet &
Resnet-50 \\
\hline
Swin-Transformer & Adam , 1e-1 & 20 & ImageNet & Swin-Unet \\
\hline
DeepLab Resnet & Adam , 7e-2 & 25 &
ImageNet & Resnet-101 \\
\hline
DeepLab Xception & Adam , 7e-2 & 25 &
ImageNet & Xceeption \\
\hline
Pix2Pix U-Net & Adam , 1e-3 & 30 &
ImageNet & VGG-16 \\
\hline
Pix2Pix FPN & Adam , 1e-3 & 30 & 
ImageNet & Resnet-50 \\
\hline

\end{tabular}
\end{center}
\end{table}
\setlength{\tabcolsep}{1.4pt}

\section{Dataset details}
\label{sec:experimental-settings}
The dataset published here is the largest, most diverse and consistent crack segmentation dataset constructed so far. It contains 9255 images that combine different smaller open source datasets. It consists of 10 sub datasets preprocessed and resized to 400x400 namely, Crack500, Deepcrack, Sdnet, Cracktree, Gaps, Volker, Rissbilder, Noncrack, Masonry and Ceramic. \\
The dataset has been made publicly available and can be found at Harvard Database: 
\href{https://doi.org/10.7910/DVN/EGIEBY}{https://doi.org/10.7910/DVN/EGIEBY}.

\section{Deployment of Segmentation Pipeline using Stitching}
\label{sec:stitching}
\begin{figure}[!htp]
    \centering
    \includegraphics[width=\textwidth]{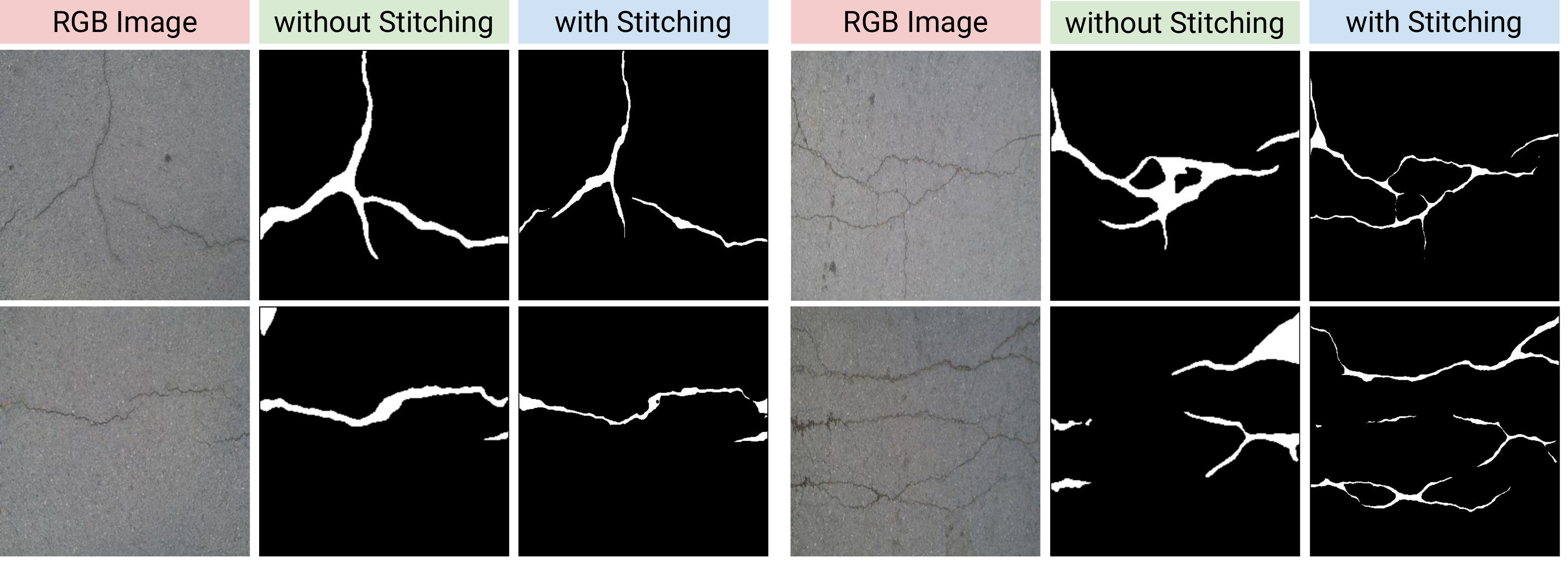}
    \caption{Transformation in Predictions due to Stitching. Model Used: Pix2Pix FPN}
    \label{fig:stitching}
\end{figure}

Inference on images with resolution higher than the input size of the model can be obtained by interpolation/ resizing but this leads to sub-par performance and artifacts in the predictions. Here, we propose an improved technique of `stitching' together the inference on crops of smaller size to produce the final result. In this technique, we divide the image into multiple subimages such that their size is equal to that of the size that the deep learning network has been trained on. Later the predictions are combined such that their spatial locations are consistent to the original input image. Using this technique enables us to by-pass the need of YOLO for detecting crops and resizing them. The qualitative comparison of `stitching' and the previous pipeline is presented in Fig. \ref{fig:stitching}. 

\end{document}